\documentclass[letterpaper]{article}
\usepackage[preprint]{aaai2027}
\usepackage[hyphens]{url}
\usepackage{graphicx}
\usepackage{natbib}
\usepackage{caption}
\usepackage{amsmath,amssymb}
\usepackage{booktabs}
\usepackage{xcolor}

\definecolor{okgreen}{RGB}{0,120,70}
\definecolor{badred}{RGB}{178,34,34}
\definecolor{cardgray}{RGB}{242,242,242}
\definecolor{staleblue}{RGB}{225,235,245}
\title{Stale-Document Poisoning:\\
When Outdated Retrieval Overrides Correct Model Answers}
\author{
Md Shamim Ahmed\textsuperscript{\rm 1}\corresponding,
Lukas Galke Poech\textsuperscript{\rm 1},
Richard R\"{o}ttger\textsuperscript{\rm 1}
}
\affiliations{
\textsuperscript{\rm 1}University of Southern Denmark\\
shamim@imada.sdu.dk
}

\begin{document}
\maketitle
\begin{abstract}

Retrieval-augmented generation (RAG) is often used to address outdated knowledge by providing external evidence. But retrieval helps only when that evidence is still valid. We identify a temporal alignment failure, \emph{stale-document poisoning}, in which outdated evidence makes a model wrong despite answering correctly without retrieval.

We construct a benchmark of 317 verified \emph{knowledge reversals} across medicine, law, software, and platform policy, grounded in dated official sources. Across 12 models, recent medical reversals are harder than long-established ones. More importantly, outdated retrieval flips 30\% of Llama and 37\% of Qwen answers even without instructions to trust the document; explicit follow instructions raise these rates to 66\% and 75\%. Across four open models and four domains, poisoning ranges from 17--91\%, while matched up-to-date evidence is followed in 97--100\% of trials.

To isolate temporal applicability, we keep the historical evidence unchanged across 50 reversals and vary only the evaluation date. A clear pattern emerges: dates alone produce only modest adaptation, but when models are explicitly told when the old evidence stops applying, the larger models switch to the appropriate answer almost perfectly. Causal interventions confirm that this validity information directly shapes the final decision. The same internal components also support broader comparison tasks, suggesting that temporal applicability can recruit a general reasoning mechanism used for other comparisons.

Finally, a fixed recency-aware hybrid re-ranker reduces poisoning by 4.6--10.0 points when dates are accurate, with gains that depend on reliable temporal metadata. Reliable RAG therefore requires \emph{selective trust}: models must determine not only what retrieved evidence says, but whether it still applies.

\end{abstract}

\section{Introduction}

Medical guidance can change faster than models can be retrained. Routine beta-blockade after myocardial infarction with preserved ejection fraction, vitamin D with calcium for fracture prevention, and fixed-duration antibiotic regimens have each been standard recommendations that were later withdrawn or narrowed. Legal doctrine can likewise reverse: in 2024, the U.S.\ Supreme Court's decision in \emph{Loper Bright Enterprises v.\ Raimondo} overruled \emph{Chevron U.S.A. Inc.\ v.\ Natural Resources Defense Council}, ending a four-decade doctrine of judicial deference to federal agencies. Software knowledge changes when APIs are deprecated or replaced. Models trained before such changes may therefore retain recommendations that were once correct but are no longer applicable \citep{lazaridou2021mind,vu2024freshllms,cheng2024dated}.

Retrieval-augmented generation (RAG) is a natural response to this problem: rather than relying only on model weights, the system retrieves external evidence at inference time. But retrieval introduces a second temporal problem. The retrieved evidence may itself be outdated. Prior work shows that incorrect or counterfactual context can override an otherwise-correct answer \citep{longpre2021entity,xie2024adaptive,wu2024clasheval}. An outdated document can create the same conflict without adversarial fabrication: evidence that was once valid may now support a superseded answer. We call this failure \textbf{stale-document poisoning}. Unlike adversarial poisoning and indirect prompt injection \citep{zou2025poisonedrag,greshake2023not}, the evidence need not be fabricated or maliciously constructed; it becomes harmful simply because the conditions under which it was valid have changed.

The deeper problem is \textbf{temporal applicability}. Knowing when a document was written is not enough. A model must determine whether the recommendation expressed by that document still applies at the time of the question, and that judgment must influence the final answer. The same historical evidence can therefore be appropriate at one evaluation date and stale at another even though its text is unchanged. This distinguishes temporal applicability from generic conflict between parametric knowledge and retrieved context: the relevant question is not merely which source the model follows, but whether its deference changes when the evidence crosses a verified validity boundary.

Viewed as an alignment problem, this is a failure of \textbf{selective epistemic trust}. A reliable model should defer to retrieved evidence when that evidence remains applicable, not simply because it appears in context or because the prompt instructs the model to follow it. Our matched experiments test this distinction directly. Outdated evidence can overturn otherwise-correct answers even under neutral retrieval, explicit document-following instructions amplify the effect, and matched up-to-date evidence is followed almost universally. Thus, the failure is not a general inability to use retrieval; it is an inability to reliably condition reliance on whether retrieved evidence should still govern the answer.

\begin{figure*}[t]
\centering
\includegraphics[width=\textwidth]{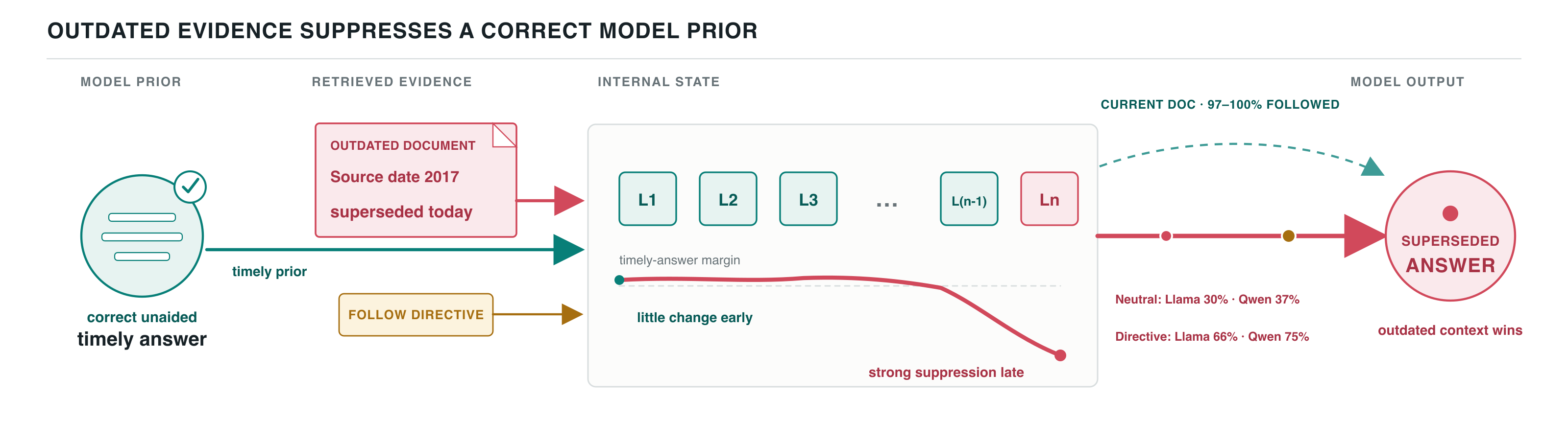}
\caption{Retrieving outdated evidence can overturn an answer that the same model gets right without retrieval. The timely-answer margin changes little in early layers but drops sharply near the output. Even neutral retrieval causes substantial poisoning, and an explicit follow directive amplifies the effect. In contrast, matched up-to-date evidence is followed almost universally, revealing a failure of selective trust rather than a general failure to use retrieval.}
\label{fig}
\end{figure*}

We study this failure at three levels. First, we measure \emph{behavioral poisoning}: how often does outdated evidence overturn an answer that the same model gives correctly without retrieval? Second, we test whether models can tell when the same evidence is still valid and when it has become outdated. Across 50 source-verified reversals, we hold the historical evidence fixed and vary only the evaluation date, asking whether models change their reliance on the same evidence when it moves from valid to superseded. Models respond only modestly to date changes alone, but change their answers sharply when the evidence’s validity boundary is made explicit. This shows that knowing the dates is not enough; models must also determine whether the evidence is still valid at the time of the question.

Third, we examine how temporal applicability affects the model's internal decision process. Causal interventions show that information about when evidence applies can directly change the model's final answer, while matched control interventions have almost no effect. Further analysis points to attention as an important early contributor, with the same components also supporting more general comparison tasks. Together, these results suggest that models already have internal mechanisms capable of using explicit validity information; the key challenge is making that information reliably guide how much they trust retrieved evidence.

Reliable RAG needs more than simply favoring newer documents. We therefore test a fixed hybrid re-ranker that combines semantic relevance, document recency, and supersession cues. It consistently reduces poisoning when date metadata are reliable, showing that temporal information can improve retrieval before generation. Its effectiveness, however, depends on metadata quality. The broader solution is therefore validity-aware evidence arbitration: retrieval systems should determine not only which evidence is relevant, but whether it still applies.

Prior work shows that models can become outdated, that retrieved evidence can conflict with what they already know, and that models often defer strongly to external information (Section~\ref{sec:related}). We ask a more specific question: \emph{when a model already has the correct answer, can it tell whether retrieved evidence should still be trusted?} We test this by measuring poisoning only when retrieval overturns an otherwise-correct answer, comparing matched current and superseded evidence, and then holding the evidence completely fixed while changing only the evaluation date. This isolates whether model trust actually tracks the validity of the retrieved evidence, rather than simply whether the model follows context.

Our primary evaluation covers 12 models: six API-accessed models and six open-weight models spanning 4B to 72B parameters. Cross-domain behavioral experiments cover Qwen2.5 and Llama-3.1 models at both small and large scales; temporal-applicability experiments include Qwen2.5-7B/72B and Llama-3.1-8B/70B, with causal analyses focused on the 72B and 70B models. We additionally include a GPT-5.5 poisoning stress test.

Our main contributions are:
\begin{enumerate}

\item We introduce a source-verified benchmark of 317 knowledge reversals across medicine, law, software, and platform policy. Across models and domains, a single outdated evidence item can overturn an answer the same model gets right without retrieval.

\item We show that this failure reflects \textbf{selective epistemic trust}: models readily follow valid evidence, but do not reliably resist outdated evidence. Poisoning occurs even under neutral retrieval and increases sharply under explicit follow instructions.

\item Using 50 source-verified temporal transitions, we hold the evidence fixed and vary only the evaluation date. Dates alone produce little change, while explicit information about when the evidence stops applying enables large models to switch to the correct answer almost perfectly.

\item Causal interventions show that explicit validity information can directly change the model's answer. The same internal components also support broader comparison tasks, suggesting that temporal applicability draws on a general comparison process rather than a dedicated temporal mechanism.

\item We show that retrieval-side mitigation depends on trustworthy temporal metadata. A fixed recency-aware hybrid re-ranker reduces poisoning when dates are reliable, motivating validity-aware evidence arbitration rather than recency alone.

\end{enumerate}

\section{Benchmark and Setup}
\textbf{Problem formulation and estimands.}
For item $i$, let $q_i$ be the question, $d_i$ authentic historical evidence
supporting the formerly correct answer $y_i^{\mathrm{old}}$, and $y_i^*$ the
timely answer. For the temporal-applicability subset, let $I_i$ denote the
verified period in which $d_i$ remains valid, and define
$g_i(t)=\mathbb{1}[t\in I_i]$. Let $R_m(v)$ denote model $m$'s old-answer rate
when $g_i(t)=v$.

For model $m$, let $E_m$ contain the items answered correctly without
retrieval. We measure two quantities:
\begin{equation}
\begin{aligned}
P_m &=
\frac{1}{|E_m|}\sum_{i\in E_m}
\mathbb{1}[\hat y_m(q_i,d_i,t_{\mathrm{stale}})\ne y_i^*],\\
G_m &= R_m(1)-R_m(0).
\end{aligned}
\label{eq:estimands}
\end{equation}

$P_m$ measures how often outdated retrieval overturns an otherwise-correct
answer. $G_m$ measures how strongly the model changes its reliance on the
\emph{same evidence} when that evidence moves from valid to superseded; larger
values indicate better temporal-applicability discrimination. No malicious
author or fabricated document is required: the failure arises when a model
continues to trust genuine evidence after it no longer applies.

\textbf{Knowledge reversals.}
Each benchmark item contains a question, a \emph{timely answer} that is correct at evaluation time, and a \emph{superseded answer} that was once correct but is no longer valid. An \emph{up-to-date document} supports the timely answer, while an \emph{outdated document} supports the superseded answer. We count a response as \emph{poisoned} only when the model answers correctly without retrieval but becomes wrong after receiving an outdated document. This isolates failures caused by retrieval rather than errors the model already makes on its own. The benchmark contains 317 reversals: 87 in medicine, 100 in law, 60 in software/API, and 70 in platform policy. We use the medical subset for the 12-model recency analysis and all four domains for the open-model cross-domain analysis. All reported experiments use the same fixed, versioned benchmark snapshot so that later updates to the corpus cannot change the reported results. Internal-state and intervention analyses use clearly reported subsets that satisfy the requirements of each experiment. Automated checks verify source provenance, temporal validity, duplicate items, and archive integrity; the nonmedical corpus passed all checks. For medicine, 39 of 87 official source pages were successfully archived automatically; for the remaining pages, we retain the source URLs and capture status.

\textbf{Document construction.} Each evidence item is derived from a dated official source supporting either the timely or superseded answer. Up-to-date and outdated documents use the same format, differing only in the recommendation they support and whether that recommendation is still valid.

\textbf{Temporal-applicability control.}
We construct a source-verified subset of 50 reversals for which official sources
clearly identify when an earlier recommendation stopped applying. For each item,
we present the same historical evidence at two evaluation dates: once while the
evidence is still valid and once after it has been superseded. The evidence,
question, answer options, and instructions remain identical; only the evaluation
date changes. We compare this date-only setting with two conditions that
explicitly state when the evidence stops applying, either in prose or in a small
table. Qwen2.5-7B/72B and Llama-3.1-8B/70B judge whether the evidence applies
and choose between two randomized answer options, giving a direct behavioral
measure without free-text judging. We define the \emph{applicability gap} as
the old-answer rate while the evidence is valid minus its rate after the
evidence becomes outdated.

\textbf{Causal interventions.}
We next test whether the internal information associated with the evaluation
date can directly change the model's answer. For each large model, we analyze
up to 20 items that reproduce the expected valid-date and stale-date answer
preferences under the answer-logit measure. We swap the internal state at the
evaluation-date position between the paired prompts in both directions.
Self-patches serve as a sham control, while identical patches around the
unchanged source date test whether the effect is specific to the evaluation
date. We then examine where this intervention begins to affect the final
decision and separate the contributions of attention and MLP layers. Candidate
attention heads are identified on 10 discovery items, frozen, and tested on a
separate 10-item confirmation set. Finally, we test the same heads on ordinary
date comparisons and non-temporal threshold tasks to determine whether they are
specific to temporal applicability or support comparison more generally.

\textbf{Fixed evaluation, refreshable corpus.} We evaluate all models on a fixed benchmark snapshot with explicit date annotations so that results remain reproducible over time. Future releases may add newly verified reversals, but they will not modify the evaluation set used in this paper. If guidance changes again, we record the new source and effective date in a later corpus version while preserving the version evaluated here. Versioned snapshots and a changelog preserve past evaluations while allowing the corpus to incorporate new reversals.

\textbf{Recency control.}
Recent reversals may be harder either because the knowledge changed recently or because the questions themselves are more difficult. We therefore compare each model on settled medical reversals, which predate training by years, and recent reversals occurring close to or after training. Models perform near ceiling on settled items but substantially worse on recent ones, supporting a recency effect. Because the two sets contain different questions, residual difficulty differences cannot be ruled out, and this control is limited to medicine. To our knowledge, prior work has not tested this same-model comparison on real, source-verified medical reversals \citep[e.g.,][]{park2025chroknowledge,vladika2025facts,guan2026tempomed}.

\textbf{Models.}
We evaluate 12 models: six API-accessed frontier models (GPT-4o, GPT-4.1,
Claude-Opus-4.5, Claude-Sonnet-4.6, DeepSeek-V3, and DeepSeek-R1) and six
open-weight models (Qwen2.5-7B/72B \citep{qwen2024qwen25},
Llama-3.1-8B/70B \citep{grattafiori2024llama3}, and MedGemma-4B/27B
\citep{sellergren2025medgemma}). Llama 3.1 has a reported December 2023
knowledge cutoff; exact cutoffs are unavailable for Qwen2.5 and MedGemma.
Model identifiers, revisions, and run dates appear in the supplement.
Internal analyses use Qwen2.5-7B and Llama-3.1-8B, while causal
temporal-applicability experiments use Qwen2.5-72B and Llama-3.1-70B.

\textbf{Automatic judge.}
We use an automatic judge to classify each free-text response as
\textbf{timely, superseded, or unclear} by comparing it with the item's timely
and superseded reference answers. We validate this judge in two independent
ways. First, a second judge from a different model family independently
re-evaluates 840 stored medical responses from seven models. The two judges
show near-perfect agreement (Cohen's $\kappa=0.93$).

\textbf{Human validation.}
We then evaluate the automatic judge against blinded human annotations. Two
external PhD students independently label a stratified sample of 200 responses,
covering all four domains, evaluated model families, and both unaided and
outdated-document conditions. Their initial agreement is 76.0%
($\kappa=0.621$), and a separate blinded adjudicator resolves the 48
disagreements. Against these final human labels, the automatic judge achieves
92.5\% accuracy (95\% CI, 88.9--95.9\%) and a macro-F1 of 0.894
(95\% CI, 0.836--0.941). Confidence intervals are estimated with 10,000
bootstrap samples clustered by the 152 underlying benchmark items.
Materials contained no patient records or personal or sensitive data, and
annotators were not asked to provide domain advice.

\section{Results}
Our results establish three linked findings: outdated evidence overturns correct answers across models and domains; dates alone provide little control over whether that evidence is followed; and explicit applicability information causally reaches the answer through an attention-led comparison process. We then examine instruction pressure and retrieval-side mitigation.
\subsection{Staleness and Poisoning}

\begin{figure*}[t]
\centering
\includegraphics[width=\textwidth]{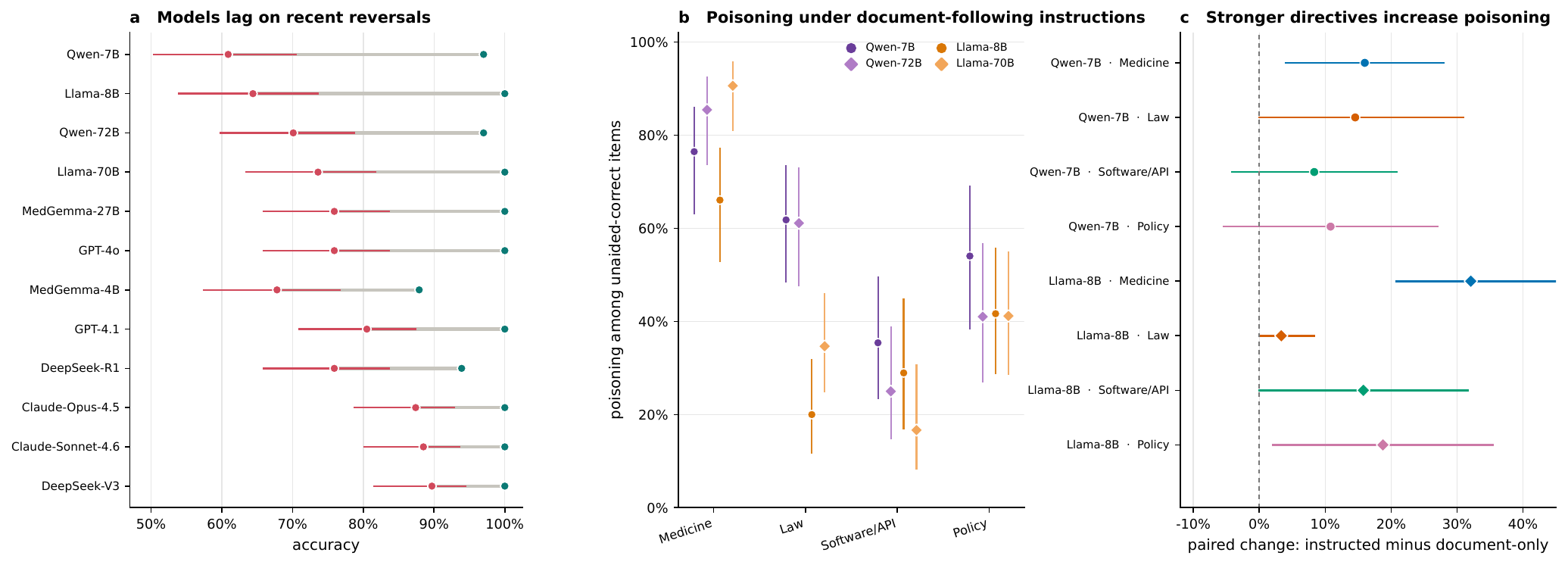}
\caption{\textbf{Stale knowledge is harder to answer, outdated retrieval overturns correct answers, and follow instructions increase poisoning.}
(a) Medical accuracy is near ceiling on settled reversals but lower on recent ones; intervals are Wilson 95\% CIs.
(b) A single outdated document overturns otherwise-correct answers across all four domains; points show poisoning rates with Wilson 95\% CIs.
(c) Explicit follow instructions increase poisoning relative to document-only retrieval; bars show paired-bootstrap 95\% CIs.}
\label{fig:phenomenon}
\end{figure*}

\textbf{Models lag on recent reversals.} All twelve models perform at or near ceiling on settled reversals, but accuracy drops substantially on recent ones—for example, to $0.61$ for Qwen-7B, $0.64$ for Llama-8B, $0.76$ for GPT-4o, and $0.87$ for Claude-Opus-4.5 (Fig.~\ref{fig:phenomenon}a). The gap remains significant for 10 of 12 models after Benjamini--Hochberg correction. This consistent within-model pattern shows that recently changed knowledge is markedly harder, although some difference in item difficulty may also contribute.

\textbf{One outdated document can overturn correct answers across every domain.}
Under the instructed condition, poisoning is highest in medicine
($0.91$ for Llama-70B and $0.85$ for Qwen-72B), but remains substantial
in law and policy (Fig.~\ref{fig:phenomenon}b). Software is more resistant,
with poisoning rates of $0.17$--$0.35$. The effect appears in every
model--domain combination and does not disappear with scale. A full
small-versus-large model analysis finds no systematic scale effect
(supplementary material).

\textbf{High unaided accuracy does not protect against poisoning.}
GPT-5.5 answers $0.92$--$1.00$ of the evaluated questions correctly without
retrieval, yet a single outdated document overturns $74/83$ previously correct
medical answers. The nonmedical samples are smaller and are therefore treated
as stress tests rather than prevalence estimates. Thus, stale-document
poisoning is not simply a consequence of weak models or poor underlying
knowledge: even a highly accurate model can abandon a correct answer when
presented with superseded evidence.

\subsection{Dates Alone Do Not Reliably Determine Temporal Applicability}
\label{sec:applicability}

The poisoning experiment shows that outdated evidence can overturn an answer
that the same model gives correctly without retrieval. However, its up-to-date
and outdated conditions necessarily support different recommendations. We
therefore isolate temporal applicability from evidence content. Across 50
source-verified reversals with explicit applicability boundaries, each model
sees the same historical evidence at two evaluation dates: once while that
evidence remains valid and once after it has been superseded. Within each pair,
the evidence, question, answer options, and instructions are identical; only
the evaluation date changes.

\begin{figure*}[t]
\centering
\includegraphics[width=\textwidth]{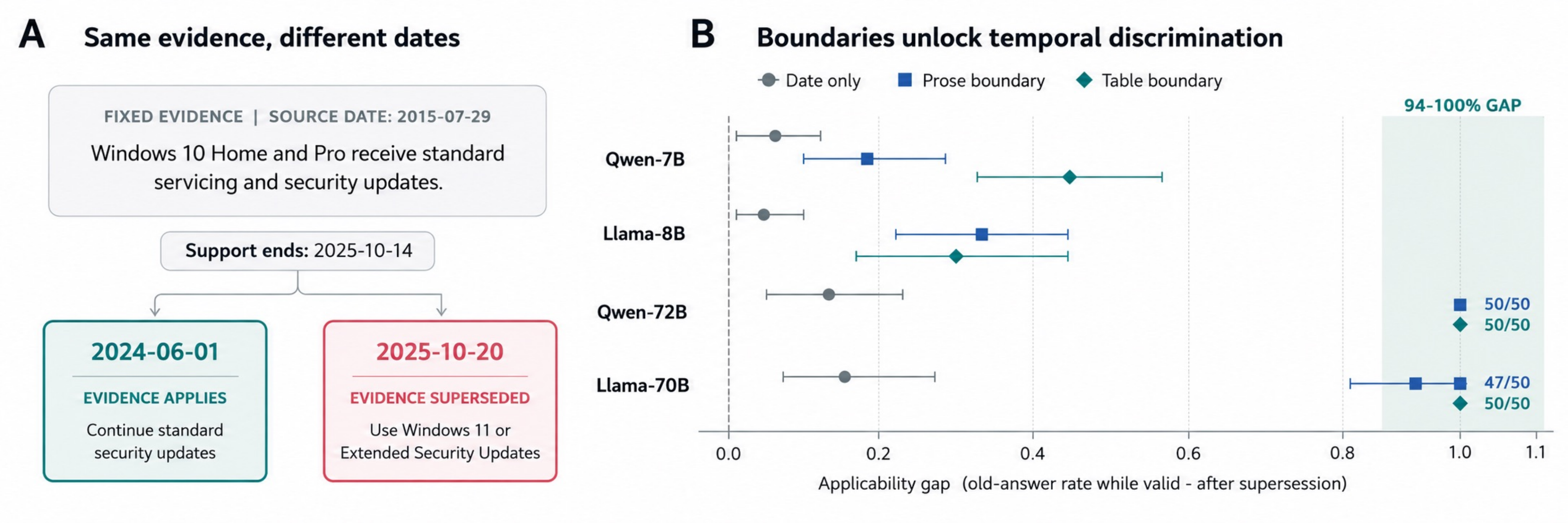}
\caption{\textbf{Dates alone provide little control; explicit validity boundaries unlock temporal discrimination in large models.}
(a) Historical evidence, question, options, and instructions are fixed; only whether the evaluation date precedes or follows the verified boundary changes.
(b) Gaps are the old-answer rate while valid minus after supersession; whiskers are item-clustered bootstrap 95\% CIs. Annotations show large-model transitions.}
\label{fig:applicability}
\end{figure*}

Dates alone produce only small changes in evidence use
(Fig.~\ref{fig:applicability}). In contrast, Qwen-72B makes all 50 required
old-to-current transitions when the validity boundary is explicit in either
format. Llama-70B makes 47/50 transitions with a prose boundary and 50/50 with
a tabular boundary. Smaller models improve more modestly, showing that the
effect is not explained simply by the change in response format.

These results distinguish having temporal information from using it correctly. Large models can follow a clear validity rule, but dates alone rarely lead them to the same decision. Explicit validity boundaries therefore reveal a strong underlying capability and provide a direct way to test whether models understand when evidence still applies.

\subsection{Explicit Applicability Information Causally Shapes the Answer}
\label{sec:causal}

We next ask whether information about when evidence applies can directly change the model's answer. To test this, we swap the internal state at the evaluation-date position between paired prompts in which the same evidence is either still valid or already outdated.

The effect is large and highly specific. Moving the valid-date state into the stale prompt transfers most of the answer preference: $23.08$ logits for Qwen-72B (95\% CI $[21.06,25.53]$) and $7.82$ for Llama-70B ($[6.92,8.65]$). Reversing the patch shifts the answer in the opposite direction. In contrast, patches at the unchanged source date move the margin by only $.04$ and $.03$, and self-patches have essentially no effect. This shows that the effect is tied specifically to the evaluation-date information.

We then trace where this effect appears in the final decision. In both models, it becomes increasingly visible in later layers. For Llama-70B, the effect reaches $1.21$ logits at layer 31 and $5.86$ at layer 39, recovering 75\% of the transferred preference. For Qwen-72B, it reaches $3.66$ at layer 55 and grows to 80\% of the transferred effect by layer 63. Matched controls remain near zero throughout.

Attention is the main early contributor. At the first layer with a clear effect, attention contributes $0.91$ logits in Llama-70B compared with $-0.17$ from the MLP; for Qwen-72B, the corresponding values are $2.34$ and $-0.04$. We identify candidate attention heads using 10 discovery items, freeze the selection, and test them on a separate 10-item confirmation set. Eight of ten Qwen heads and seven of fifteen Llama heads remain significant after Holm correction, while matched-control effects stay close to zero.

These heads also contribute to ordinary date comparisons and non-temporal threshold decisions. They therefore appear to support a broader comparison process rather than a mechanism dedicated only to time. Together, the results show that large models can internally use explicit information about when evidence applies, and that this information can causally shape the final answer.

\subsection{Instruction Pressure Amplifies Poisoning}
\label{sec:instruction}

\textbf{The effect is not caused by the ``most current'' label.}
In a $2\times2$ experiment, we vary both the document header and whether the
model is explicitly told to follow the retrieved evidence. With a neutral
header, the follow instruction raises poisoning from $0.37$ to $0.75$ in Qwen
($n=51$; OR $4.92$, 95\% CI $[2.59,9.34]$) and from $0.30$ to $0.66$ in
Llama ($n=53$; OR $4.50$, 95\% CI $[2.46,8.23]$). In both models, 19 items
become incorrect and none improve (exact McNemar
$p=3.8\times10^{-6}$). Once the follow instruction is present, calling the
document ``most current'' adds no detectable effect.

\textbf{Instruction pressure amplifies poisoning across domains.}
In all eight model--domain comparisons, explicitly telling the model to follow
outdated evidence causes more poisoning than presenting the evidence alone.
Across the nonmedical domains, Qwen-7B shows 28 deteriorations versus
12 improvements ($p=0.0166$), while Llama-8B shows 23 versus 6
($p=0.0023$). Thus, outdated evidence can already overturn correct answers
under neutral retrieval, and explicit instructions to follow it make the
failure substantially worse.

Recognizing that evidence may be outdated is also not enough. In a separate
experiment, models sometimes identify evidence as superseded or uncertain but
still follow it when answering (supplementary material).
\subsection{Retrieval-Side Mitigation Depends on Metadata Quality}

We next test whether stale evidence can be filtered before it reaches the model.
Each retrieval set contains one up-to-date and two outdated documents, and a
fixed hybrid re-ranker combines semantic relevance with document recency.

When dates are accurate, re-ranking reduces downstream poisoning by
$4.6$--$10.0$ percentage points across Qwen-7B and GPT-4o in medicine and law,
with three of four comparisons significant at $p<0.05$. The gains become small
when dates are unavailable, while incorrect dates can favor outdated evidence.
Thus, temporal metadata can improve retrieval when it is trustworthy, but
recency alone cannot determine whether evidence still applies.

Full retrieval results and additional internal-intervention experiments appear
in the supplementary material.

\section{Related Work}
\label{sec:related}

We situate stale-document poisoning at the intersection of temporal knowledge, RAG knowledge conflict, internal model control, and deference to retrieved evidence.

\textbf{Temporal staleness and evolving knowledge.}
Prior work on time-sensitive QA
\citep{dhingra2022time,chen2021dataset,liska2022streamingqa,kasai2023realtime}
and superseded knowledge
\citep{lazaridou2021mind,vu2024freshllms}
shows that language models often lag behind changing facts, with effective knowledge cutoffs sometimes preceding reported ones
\citep{cheng2024dated}. Related benchmarks study evolving knowledge across domains
\citep{park2025chroknowledge,zhu2025evolvebench}, changing medical guidance
\citep{vladika2025facts,guan2026tempomed}, and overruled legal precedent
\citep{zhang2025precedent}.

Our question is different: even when a model already knows the current answer, can it determine whether retrieved evidence still applies? We isolate this directly by holding the retrieved evidence byte-identical and changing only the evaluation date. This separates temporal applicability from both outdated model knowledge and differences in document content.

\textbf{Knowledge conflict and RAG poisoning.}
Prior work shows that retrieved context can conflict with a model's internal
knowledge, using entity substitutions \citep{longpre2021entity}, coherent
counter-evidence \citep{xie2024adaptive}, benchmarked conflicts
\citep{wu2024clasheval,marjanovic2024dynamicqa}, and real documents
\citep{kortukov2024studying}; see \citet{xu2024knowledge} for a review.
Recent methods address these conflicts through source balancing
\citep{li2026balance}, context-compliance analysis
\citep{chen2026contextcompliance}, knowledge selection
\citep{zhang2025knowpo}, explicit conflict signals
\citep{ye2026seeing}, and adaptation of retrieved evidence
\citep{wang2026accommodate}.

Our setting is more specific: we test whether a model's deference changes with
the temporal validity of real documents while holding the item and instruction
condition fixed. PoisonedRAG \citep{zou2025poisonedrag} and indirect prompt
injection \citep{greshake2023not} provide the adversarial framing, but our
documents are not maliciously constructed: they were once correct and become
harmful only because they are outdated.

\textbf{Internal signals and control.}
Prior work shows that factual recall and context--memory conflict leave
detectable internal signatures
\citep{geva2023dissecting,yu2023characterizing,ortu2024competition}, while
causal localization does not necessarily identify where an intervention will
most effectively change behavior \citep{hase2023does}. Causal tracing and activation patching
go further by intervening on internal states to test which representations
influence model outputs \citep{meng2022locating}. Related work has localized
behaviors to attention heads and modified them through head- and
activation-level interventions
\citep{jin2024cutting,minder2025controllable,arditi2024refusal}. We extend this
line of work to temporal applicability using bidirectional patching with matched
controls and a discovery--confirmation split. Our interventions track how
information supplied at the evaluation-date positions influences the final
answer and identify attention components enriched for this effect. Specificity
controls further test whether these components are specialized for temporal
applicability or participate in a broader comparison-and-decision process.

\textbf{Sycophancy and reliability.}
Post-training can increase agreement with user-provided claims
\citep{perez2023discovering,sharma2024towards}, although models can also resist
deceptive context \citep{samsami2024toobig}. Because we vary instruction
strength within each model family, we do not attribute stale-document poisoning
specifically to post-training.

\section{Discussion}

\textbf{Key findings.}
Our central result is that retrieval can make a model wrong when it was already right. A single outdated evidence item can overturn an answer that the same model gives correctly without retrieval, across models and domains and even under neutral retrieval. Explicit instructions to follow the retrieved evidence amplify this failure. In contrast, matched up-to-date evidence is followed in $97$--$100\%$ of trials. The problem is therefore not a general inability to use retrieval, but a failure of \textbf{selective epistemic trust}: models do not reliably condition their deference on whether retrieved evidence still applies.

The temporal-applicability experiment isolates this failure more directly. Holding the historical evidence, question, answer options, and instructions fixed while changing only the evaluation date produces little discrimination from dates alone. Yet when the evidence's validity boundary is made explicit, the 70B/72B models switch almost perfectly between the historical and current answers. This separates access to temporal information from the ability to use that information: capable models can apply an explicit validity rule, but do not reliably infer the same rule from ordinary date metadata.

The causal experiments show that explicit applicability information can directly shape the model's decision. Changing internal states associated with the evaluation date strongly changes the answer, while matched controls have little effect. Attention contributes early to this process, but the same components also support more general comparison tasks. This suggests that temporal applicability draws on a broader comparison mechanism rather than a specialized temporal circuit.

\textbf{Validity and scope.}
The poisoning analysis is deliberately conditional: an item counts as poisoned only when retrieval overturns an answer that the same model previously gave correctly. This isolates retrieval-induced failure from errors already present without retrieval. The automatic judge was independently validated against 200 adjudicated human labels across all four domains, achieving $92.5\%$ accuracy and a macro-F1 of $0.894$. The temporal-applicability control strengthens the identification further by avoiding free-text adjudication and varying only the evaluation date within each matched pair.

Several results nevertheless have narrower scope. The settled-versus-recent comparison is limited to medicine and uses different questions, so recency cannot be completely separated from item difficulty. The causal analyses use 20 behaviorally eligible items per large model and an answer-logit task with explicit answer options. They establish causal influence in this controlled setting, but not that the same components govern unconstrained generation or deployed RAG systems. Head-level specificity is evaluated on a 10-item held-out confirmation set per model and should therefore be interpreted as localization evidence rather than a complete circuit description.

\textbf{Generalizability.}
Stale-document poisoning appears across four domains and four open-weight models spanning two model families and scales from 7B to 72B. The medical evaluation extends to 12 models, and the GPT-5.5 stress test shows that high unaided accuracy does not eliminate the vulnerability. The temporal-applicability control covers 50 independently verified transitions and four models from the Qwen and Llama families. 
The much stronger performance of the 70B/72B models suggests that reliable applicability reasoning emerges with model capability and remains sensitive to how temporal information is presented.

\textbf{Practical implications.}
The results suggest that publication date and applicability are different kinds of metadata. Knowing when a document was written does not tell a system when its recommendation ceased to apply. Reliable retrieval systems should therefore represent temporal validity explicitly where possible---for example, through effective dates, supersession relations, or validity intervals---and models must be able to use that information when deciding whether retrieved evidence should change an answer.

Retrieval-side safeguards remain useful but incomplete. Our fixed recency-aware hybrid re-ranker reduces poisoning by $4.6$--$10.0$ percentage points when dates are accurate, but its benefit becomes small when dates are absent and can reverse when dates are wrong. Explicit validity boundaries provide a stronger behavioral signal for capable models, but they too depend on trustworthy provenance and metadata. The broader challenge is therefore \emph{validity-aware evidence arbitration}: determining not only whether evidence is relevant, but whether it remains applicable before allowing it to override an otherwise-correct answer.

This also exposes a blind spot in standard RAG evaluation. Retrieval should be evaluated in both directions: whether valid evidence can usefully correct an outdated model, and whether superseded evidence can harm a model that was already correct. More information is not automatically better; reliable RAG requires controlling which information still deserves epistemic authority.

\section{Conclusion}

Stale-document poisoning shows that retrieval can undermine knowledge a model already gets right. Across models and domains, outdated evidence can overturn correct answers, while matched up-to-date evidence is followed almost universally. The resulting challenge is not simply retrieving relevant information, but deciding whether that information still deserves to influence the answer.

Our temporal control sharpens this distinction. When the evidence is held fixed, dates alone produce little change, yet large models respond almost perfectly when told explicitly when the evidence stops applying. Causal interventions further show that this applicability information can directly shape the model's decision. Together, these results reveal an important gap: models can use temporal validity when it is made clear, but do not reliably infer and act on it from ordinary temporal cues.

Reliable RAG therefore requires \emph{selective epistemic trust}: systems must determine not only what retrieved evidence says, but whether it still applies. A promising direction is to move beyond publication dates toward explicit validity intervals, supersession relations, and models trained to reason over them. Retrieval should not merely find relevant evidence; it should help decide which evidence still deserves to be trusted.

\bibliography{refs}

@inproceedings{lazaridou2021mind,
  author    = {Angeliki Lazaridou and Adhiguna Kuncoro and Elena Gribovskaya and Devang Agrawal and Adam Liska and Tayfun Terzi and Mai Gimenez and Cyprien de Masson d'Autume and Tomas Kocisky and Sebastian Ruder and Dani Yogatama and Kris Cao and Susannah Young and Phil Blunsom},
  title     = {{Mind the Gap: Assessing Temporal Generalization in Neural Language Models}},
  booktitle = {Advances in Neural Information Processing Systems},
  year      = {2021},
  eprint    = {2102.01951},
}

@inproceedings{cheng2024dated,
  author    = {Jeffrey Cheng and Marc Marone and Orion Weller and Dawn Lawrie and Daniel Khashabi and Benjamin Van Durme},
  title     = {{Dated Data: Tracing Knowledge Cutoffs in Large Language Models}},
  booktitle = {Conference on Language Modeling (COLM)},
  year      = {2024},
  eprint    = {2403.12958},
}

@article{dhingra2022time,
  author  = {Bhuwan Dhingra and Jeremy R. Cole and Julian Martin Eisenschlos and Daniel Gillick and Jacob Eisenstein and William W. Cohen},
  title   = {{Time-Aware Language Models as Temporal Knowledge Bases}},
  journal = {Transactions of the Association for Computational Linguistics},
  year    = {2022},
  eprint  = {2106.15110},
}

@inproceedings{chen2021dataset,
  author    = {Wenhu Chen and Xinyi Wang and William Yang Wang},
  title     = {{A Dataset for Answering Time-Sensitive Questions}},
  booktitle = {NeurIPS Datasets and Benchmarks Track},
  year      = {2021},
  eprint    = {2108.06314},
}

@inproceedings{liska2022streamingqa,
  author    = {Adam Li{\v s}ka and Tom{\'a}{\v s} Ko{\v c}isk{\'y} and Elena Gribovskaya and Tayfun Terzi and Eren Sezener and Devang Agrawal and Cyprien de Masson d'Autume and Tim Scholtes and Manzil Zaheer and Susannah Young and Ellen Gilsenan-McMahon and Sophia Austin and Phil Blunsom and Angeliki Lazaridou},
  title     = {{StreamingQA: A Benchmark for Adaptation to New Knowledge over Time in Question Answering Models}},
  booktitle = {International Conference on Machine Learning},
  year      = {2022},
  eprint    = {2205.11388},
}

@inproceedings{kasai2023realtime,
  author    = {Jungo Kasai and Keisuke Sakaguchi and Yoichi Takahashi and Ronan Le Bras and Akari Asai and Xinyan Yu and Dragomir Radev and Noah A. Smith and Yejin Choi and Kentaro Inui},
  title     = {{RealTime QA: What's the Answer Right Now?}},
  booktitle = {NeurIPS Datasets and Benchmarks Track},
  year      = {2023},
  eprint    = {2207.13332},
}

@inproceedings{vu2024freshllms,
  author    = {Tu Vu and Mohit Iyyer and Xuezhi Wang and Noah Constant and Jerry Wei and Jason Wei and Chris Tar and Yun-Hsuan Sung and Denny Zhou and Quoc Le and Thang Luong},
  title     = {{FreshLLMs: Refreshing Large Language Models with Search Engine Augmentation}},
  booktitle = {Findings of the Association for Computational Linguistics: ACL 2024},
  year      = {2024},
  eprint    = {2310.03214},
}

@inproceedings{park2025chroknowledge,
  author    = {Yein Park and Chanwoong Yoon and Jungwoo Park and Donghyeon Lee and Minbyul Jeong and Jaewoo Kang},
  title     = {{ChroKnowledge: Unveiling Chronological Knowledge of Language Models in Multiple Domains}},
  booktitle = {International Conference on Learning Representations},
  year      = {2025},
  eprint    = {2410.09870},
}

@inproceedings{vladika2025facts,
  author    = {Juraj Vladika and Mahdi Dhaini and Florian Matthes},
  title     = {{Facts Fade Fast: Evaluating Memorization of Outdated Medical Knowledge in Large Language Models}},
  booktitle = {Findings of the Association for Computational Linguistics: EMNLP 2025},
  year      = {2025},
  eprint    = {2509.04304},
}

@inproceedings{zhang2025precedent,
  author    = {Li Zhang and Jaromir Savelka and Kevin Ashley},
  title     = {{Do LLMs Truly Understand When a Precedent Is Overruled?}},
  booktitle = {Legal Knowledge and Information Systems (JURIX)},
  year      = {2025},
  eprint    = {2510.20941},
}

@article{guan2026tempomed,
  author  = {Zihan Guan and Qiao Jin and Guangzhi Xiong and Fangyuan Chen and Mengxuan Hu and Qingyu Chen and Yifan Peng and Zhiyong Lu and Anil Vullikanti},
  title   = {{Large Language Models Lack Temporal Awareness of Medical Knowledge}},
  journal = {arXiv preprint arXiv:2605.13045},
  year    = {2026},
}

@inproceedings{longpre2021entity,
  author    = {Shayne Longpre and Kartik Perisetla and Anthony Chen and Nikhil Ramesh and Chris DuBois and Sameer Singh},
  title     = {{Entity-Based Knowledge Conflicts in Question Answering}},
  booktitle = {Proceedings of the Conference on Empirical Methods in Natural Language Processing},
  year      = {2021},
  eprint    = {2109.05052},
}

@inproceedings{xie2024adaptive,
  author    = {Jian Xie and Kai Zhang and Jiangjie Chen and Renze Lou and Yu Su},
  title     = {{Adaptive Chameleon or Stubborn Sloth: Revealing the Behavior of Large Language Models in Knowledge Conflicts}},
  booktitle = {International Conference on Learning Representations},
  year      = {2024},
  eprint    = {2305.13300},
}

@inproceedings{wu2024clasheval,
  author    = {Kevin Wu and Eric Wu and James Zou},
  title     = {{ClashEval: Quantifying the tug-of-war between an LLM's internal prior and external evidence}},
  booktitle = {NeurIPS Datasets and Benchmarks Track},
  year      = {2024},
  eprint    = {2404.10198},
}

@inproceedings{xu2024knowledge,
  author    = {Rongwu Xu and Zehan Qi and Zhijiang Guo and Cunxiang Wang and Hongru Wang and Yue Zhang and Wei Xu},
  title     = {{Knowledge Conflicts for LLMs: A Survey}},
  booktitle = {Proceedings of the Conference on Empirical Methods in Natural Language Processing},
  year      = {2024},
  eprint    = {2403.08319},
}

@inproceedings{marjanovic2024dynamicqa,
  author    = {Sara Vera Marjanovi{\'c} and Haeun Yu and Pepa Atanasova and Maria Maistro and Christina Lioma and Isabelle Augenstein},
  title     = {{DYNAMICQA: Tracing Internal Knowledge Conflicts in Language Models}},
  booktitle = {Findings of the Association for Computational Linguistics: EMNLP 2024},
  year      = {2024},
  eprint    = {2407.17023},
}

@inproceedings{kortukov2024studying,
  author    = {Evgenii Kortukov and Alexander Rubinstein and Elisa Nguyen and Seong Joon Oh},
  title     = {{Studying Large Language Model Behaviors Under Context-Memory Conflicts With Real Documents}},
  booktitle = {Conference on Language Modeling (COLM)},
  year      = {2024},
  eprint    = {2404.16032},
}

@inproceedings{zou2025poisonedrag,
  author    = {Wei Zou and Runpeng Geng and Binghui Wang and Jinyuan Jia},
  title     = {{PoisonedRAG: Knowledge Corruption Attacks to Retrieval-Augmented Generation of Large Language Models}},
  booktitle = {USENIX Security Symposium},
  year      = {2025},
  eprint    = {2402.07867},
}

@inproceedings{greshake2023not,
  author    = {Kai Greshake and Sahar Abdelnabi and Shailesh Mishra and Christoph Endres and Thorsten Holz and Mario Fritz},
  title     = {{Not what you've signed up for: Compromising Real-World LLM-Integrated Applications with Indirect Prompt Injection}},
  booktitle = {Proceedings of the 16th ACM Workshop on Artificial Intelligence and Security (AISec)},
  year      = {2023},
  eprint    = {2302.12173},
}

@inproceedings{geva2023dissecting,
  author    = {Mor Geva and Jasmijn Bastings and Katja Filippova and Amir Globerson},
  title     = {{Dissecting Recall of Factual Associations in Auto-Regressive Language Models}},
  booktitle = {Proceedings of the Conference on Empirical Methods in Natural Language Processing},
  year      = {2023},
  eprint    = {2304.14767},
}

@inproceedings{hase2023does,
  author    = {Peter Hase and Mohit Bansal and Been Kim and Asma Ghandeharioun},
  title     = {{Does Localization Inform Editing? Surprising Differences in Causality-Based Localization vs. Knowledge Editing in Language Models}},
  booktitle = {Advances in Neural Information Processing Systems},
  year      = {2023},
  eprint    = {2301.04213},
}

@inproceedings{yu2023characterizing,
  author    = {Qinan Yu and Jack Merullo and Ellie Pavlick},
  title     = {{Characterizing Mechanisms for Factual Recall in Language Models}},
  booktitle = {Proceedings of the Conference on Empirical Methods in Natural Language Processing},
  year      = {2023},
  eprint    = {2310.15910},
}

@inproceedings{ortu2024competition,
  author    = {Francesco Ortu and Zhijing Jin and Diego Doimo and Mrinmaya Sachan and Alberto Cazzaniga and Bernhard Sch{\"o}lkopf},
  title     = {{Competition of Mechanisms: Tracing How Language Models Handle Facts and Counterfactuals}},
  booktitle = {Proceedings of the Annual Meeting of the Association for Computational Linguistics},
  year      = {2024},
  eprint    = {2402.11655},
}

@inproceedings{jin2024cutting,
  author    = {Zhuoran Jin and Pengfei Cao and Hongbang Yuan and Yubo Chen and Jiexin Xu and Huaijun Li and Xiaojian Jiang and Kang Liu and Jun Zhao},
  title     = {{Cutting Off the Head Ends the Conflict: A Mechanism for Interpreting and Mitigating Knowledge Conflicts in Language Models}},
  booktitle = {Findings of the Association for Computational Linguistics: ACL 2024},
  year      = {2024},
  eprint    = {2402.18154},
}

@inproceedings{minder2025controllable,
  author    = {Julian Minder and Kevin Du and Niklas Stoehr and Giovanni Monea and Chris Wendler and Robert West and Ryan Cotterell},
  title     = {{Controllable Context Sensitivity and the Knob Behind It}},
  booktitle = {International Conference on Learning Representations},
  year      = {2025},
  eprint    = {2411.07404},
}

@inproceedings{meng2022locating,
  author    = {Kevin Meng and David Bau and Alex Andonian and Yonatan Belinkov},
  title     = {{Locating and Editing Factual Associations in GPT}},
  booktitle = {Advances in Neural Information Processing Systems},
  year      = {2022},
  eprint    = {2202.05262},
}

@inproceedings{arditi2024refusal,
  author    = {Andy Arditi and Oscar Obeso and Aaquib Syed and Daniel Paleka and Nina Panickssery and Wes Gurnee and Neel Nanda},
  title     = {{Refusal in Language Models Is Mediated by a Single Direction}},
  booktitle = {Advances in Neural Information Processing Systems},
  year      = {2024},
  eprint    = {2406.11717},
}

@inproceedings{perez2023discovering,
  author    = {Ethan Perez and others},
  title     = {{Discovering Language Model Behaviors with Model-Written Evaluations}},
  booktitle = {Findings of the Association for Computational Linguistics: ACL 2023},
  year      = {2023},
  eprint    = {2212.09251},
}

@article{samsami2024toobig,
  author  = {Mohammad Reza Samsami and Mats Leon Richter and Juan Rodriguez and Megh Thakkar and Sarath Chandar and Maxime Gasse},
  title   = {{Too Big to Fool: Resisting Deception in Language Models}},
  journal = {arXiv preprint arXiv:2412.10558},
  year    = {2024},
}

@inproceedings{sharma2024towards,
  author    = {Mrinank Sharma and Meg Tong and Tomasz Korbak and David Duvenaud and Amanda Askell and Samuel R. Bowman and Newton Cheng and Esin Durmus and Zac Hatfield-Dodds and Scott R. Johnston and Shauna Kravec and Timothy Maxwell and Sam McCandlish and Kamal Ndousse and Oliver Rausch and Nicholas Schiefer and Da Yan and Miranda Zhang and Ethan Perez},
  title     = {{Towards Understanding Sycophancy in Language Models}},
  booktitle = {International Conference on Learning Representations},
  year      = {2024},
  eprint    = {2310.13548},
}

@article{qwen2024qwen25,
  author  = {{Qwen Team}},
  title   = {{Qwen2.5 Technical Report}},
  journal = {arXiv preprint arXiv:2412.15115},
  year    = {2024},
}

@article{grattafiori2024llama3,
  author  = {Aaron Grattafiori and others},
  title   = {{The Llama 3 Herd of Models}},
  journal = {arXiv preprint arXiv:2407.21783},
  year    = {2024},
}

@article{sellergren2025medgemma,
  author  = {Andrew Sellergren and others},
  title   = {{MedGemma Technical Report}},
  journal = {arXiv preprint arXiv:2507.05201},
  year    = {2025},
}

@inproceedings{zhu2025evolvebench,
  author    = {Zhiyuan Zhu and Yusheng Liao and Zhe Chen and Yuhao Wang and Yunfeng Guan and Yanfeng Wang and Yu Wang},
  title     = {{EvolveBench: A Comprehensive Benchmark for Assessing Temporal Awareness in LLMs on Evolving Knowledge}},
  booktitle = {Proceedings of the 63rd Annual Meeting of the Association for Computational Linguistics (Volume 1: Long Papers)},
  pages     = {16173--16188},
  publisher = {Association for Computational Linguistics},
  year      = {2025},
  doi       = {10.18653/v1/2025.acl-long.788},
}

@inproceedings{li2026balance,
  author    = {Shuowei Li and Haoxin Li and Wenda Chu and Yi Fang},
  title     = {{How Large Language Models Balance Internal Knowledge with User and Document Assertions}},
  booktitle = {Findings of the Association for Computational Linguistics: ACL 2026},
  pages     = {25323--25346},
  publisher = {Association for Computational Linguistics},
  year      = {2026},
  doi       = {10.18653/v1/2026.findings-acl.1267},
}

@article{chen2026contextcompliance,
  author  = {Yihang Chen and Pin Qian and Su Wang and Sipeng Zhang and Huan Xu and Shuhuai Lin and Xinpeng Wei},
  title   = {{Does RAG Know When Retrieval Is Wrong? Diagnosing Context Compliance under Knowledge Conflict}},
  journal = {arXiv preprint arXiv:2605.14473},
  year    = {2026},
}

@inproceedings{zhang2025knowpo,
  author    = {Ruizhe Zhang and Yongxin Xu and Yuzhen Xiao and Runchuan Zhu and Xinke Jiang and Xu Chu and Junfeng Zhao and Yasha Wang},
  title     = {{KnowPO: Knowledge-Aware Preference Optimization for Controllable Knowledge Selection in Retrieval-Augmented Language Models}},
  booktitle = {Proceedings of the AAAI Conference on Artificial Intelligence},
  volume    = {39},
  number    = {24},
  pages     = {25895--25903},
  year      = {2025},
  doi       = {10.1609/aaai.v39i24.34783},
}

@inproceedings{ye2026seeing,
  author    = {Hua Ye and Siyuan Chen and Ziqi Zhong and Canran Xiao and Haoliang Zhang and Yuhan Wu and Fei Shen},
  title     = {{Seeing through the Conflict: Transparent Knowledge Conflict Handling in Retrieval-Augmented Generation}},
  booktitle = {Proceedings of the AAAI Conference on Artificial Intelligence},
  volume    = {40},
  number    = {40},
  pages     = {34423--34431},
  year      = {2026},
  doi       = {10.1609/aaai.v40i40.40740},
}

@inproceedings{wang2026accommodate,
  author    = {Jiatai Wang and Zhiwei Xu and Di Jin and Xuewen Yang and Tao Li},
  title     = {{Accommodate Knowledge Conflicts in Retrieval-augmented LLMs: Towards Robust Response Generation in the Wild}},
  booktitle = {Proceedings of the AAAI Conference on Artificial Intelligence},
  volume    = {40},
  number    = {39},
  pages     = {33530--33538},
  year      = {2026},
  doi       = {10.1609/aaai.v40i39.40641},
}
\clearpage
\appendix

\section{Benchmark Construction and Provenance}
\label{app:benchmark}

Each item is a knowledge reversal with a fixed timely answer, valid at
evaluation time, and a superseded answer that was formerly correct but is no
longer valid. They are stored as \texttt{current} and \texttt{abandoned} in the
release. An up-to-date document supports the timely answer; an outdated document
supports the superseded answer. The 317-item benchmark contains medicine ($n=87$), law ($n=100$),
software and APIs ($n=60$), and platform policy ($n=70$). The 12-model recency
control uses the medical subset. The open-model cross-domain grid, gradient,
scale analysis, and retrieval-defense evaluation use the frozen benchmark, with
each experiment reporting its eligible subset.

The settled control set contains 33 medical reversals that predate the open
models' release dates by years. Every benchmark item records an official source
URL. The frozen nonmedical domains add 460 archived snapshots, two per item,
and pass the corpus and archive audits with zero findings. Automated capture of
the 87 medical source pages returned 39 successful responses; the remaining
48 official URLs blocked automated retrieval or were unavailable at capture
time. The URLs and capture statuses are retained in the release.

The poisoning condition inserts one outdated document into context. The
instructed arm uses the following wording verbatim (the same string across the
behavioral, probe, and defense harnesses):

\begin{quote}\small\ttfamily
Use the dated documentation above; if it conflicts with prior knowledge,
follow the dated documentation. \{question\} Answer in two or three
sentences.
\end{quote}

The weaker arms use the same document block, verbatim ``Most current
retrieved documentation (\{year\}): Recommended approach: \{action\}.'': the
document-only arm adds no directive; the consider arm appends ``Consider the
documentation above when answering.''; the organic control drops the header
to ``Retrieved documentation from \{year\}: \{action\}.'' with no directive
(Results, Section~3.4 of the main text). The factorial adds an organic-follow cell with
that neutral header followed by the same conflict directive used in the
instructed arm. This separates the currency label from explicit deference while
holding the superseded claim and its date fixed.

For model $m$, let $E_m$ be the items answered correctly without retrieved
context, let $a^p_{mi}$ be the poisoned answer, and let $y_i^*$ be the
timely answer. We define poisoning as
\begin{equation}
  P_m=\frac{1}{|E_m|}\sum_{i\in E_m}
  \mathbb{1}[a^p_{mi}\ne y_i^*].
  \label{eq:poisoning}
\end{equation}
Thus, poisoning is conditional on unaided correctness; items answered
incorrectly without retrieval do not enter its denominator. We report the
underlying count in every
supplementary poisoning table because the eligible set varies by model and
domain.

\section{Model Snapshots and Cutoffs}
\label{app:models}

White-box analyses run on locally pinned open-weight snapshots. Full
40-character Hugging Face revisions, repository URLs, API model identifiers,
and run dates are recorded in \texttt{results/model\_revisions.json}; abbreviated
revisions appear below. Meta reports a December 2023 cutoff for Llama 3.1. The
official Qwen2.5 and MedGemma cards do not publish exact knowledge cutoffs, so
we do not assign inferred dates to those models. Hosted aliases without a
provider revision are identified as floating rather than treated as pinned.

\begin{table}[t]
\centering\small
\begin{tabular}{lc}
\toprule
Model & Recorded revision \\
\midrule
Llama-3.1-8B-Instruct & \texttt{d10aef7999a2} \\
Llama-3.1-70B-Instruct & \texttt{d50656ee28e2} \\
Qwen2.5-7B-Instruct & \texttt{a09a35458c70} \\
Qwen2.5-72B-Instruct & \texttt{495f39366efe} \\
MedGemma-4B-IT & \texttt{290cda5eeccb} \\
MedGemma-27B-Text-IT & \texttt{5b667cf2ddcf} \\
\bottomrule
\end{tabular}
\caption{Recorded open-weight revisions (abbreviated SHAs). Qwen and Llama
revisions were captured for the local runs. The MedGemma run-time SHAs were not
logged; those rows give repository heads checked on August 8, 2026. The release
records this distinction with the full revisions.}
\label{tab:snapshots}
\end{table}

The hosted medical panel used \texttt{gpt-4o}, \texttt{gpt-4.1},
\texttt{claude-sonnet-4-6}, \texttt{claude-opus-4-5}, \texttt{deepseek-chat},
and \texttt{deepseek-reasoner} on June 20--21, 2026. The frontier stress test
used the dated identifier \texttt{gpt-5.5-2026-04-23} on June 27, 2026. Exact
result artifacts are linked from the model-revision manifest.

\section{Judge Protocol and Validation}
\label{app:judge}

The automatic judge maps each free-text answer to the timely, superseded,
or unclear label defined by the item's fixed reference pair. In medicine, 840
stored answers spanning seven models were independently re-judged by a second
model family. Agreement was near-perfect (pooled Cohen's $\kappa = 0.93$;
per-model $0.86$ to $0.95$). A separate human-validation study sampled 200
responses, with 50 from each of medicine, law, code, and policy and stratified
coverage of the retained models and base and outdated-document conditions. The sample was
not enriched for automatic-judge disagreements. Two external PhD students
independently labeled all responses under blinded conditions as timely,
superseded, or unclear. Their first-pass agreement was 76.0\%
($\kappa = 0.621$). All 48 disagreements were resolved in a separate blinded
adjudication pass, producing a complete 200-response human reference set. The
automatic judge matched that set on 185 responses (92.5\% accuracy;
cluster-bootstrap 95\% CI, 88.9--95.9\%) and achieved a macro-F1 of 0.894
(95\% CI, 0.836--0.941). Confidence intervals were estimated from 10,000
bootstrap samples clustered by the 152 underlying benchmark items. Group
differences elsewhere use Fisher exact, exact McNemar, and
Cochran-Mantel-Haenszel tests, with Benjamini-Hochberg correction across the
twelve per-model staleness comparisons.

\paragraph{Ethics statement.}
The two external PhD annotators labeled only synthetic vignettes and stored
model outputs under blinding. Materials contained no patient records or
personal or sensitive data, and annotators were not asked to provide domain
advice.

\begin{table*}[t]
\centering
\scriptsize
\setlength{\tabcolsep}{4pt}
\begin{tabular}{@{}lrrrrrrrr@{}}
\toprule
Scope & $n$ & Accuracy & Accuracy 95\% CI & Macro-F1 & Macro-F1 95\% CI & F1$_{\mathrm{tim}}$ & F1$_{\mathrm{sup}}$ & F1$_{\mathrm{unc}}$ \\
\midrule
Overall  & 200 & 0.925 & [0.889, 0.959] & 0.894 & [0.836, 0.941] & 0.945 & 0.957 & 0.780 \\
Medicine &  50 & 0.960 & [0.904, 1.000] & 0.948 & [0.840, 1.000] & 0.960 & 0.974 & 0.909 \\
Law      &  50 & 0.960 & [0.900, 1.000] & 0.917 & [0.788, 1.000] & 0.952 & 1.000 & 0.800 \\
Code     &  50 & 0.940 & [0.875, 1.000] & 0.939 & [0.849, 1.000] & 0.947 & 1.000 & 0.870 \\
Policy   &  50 & 0.840 & [0.745, 0.935] & 0.768 & [0.589, 0.898] & 0.923 & 0.848 & 0.533 \\
\bottomrule
\end{tabular}
\caption{Automatic-judge performance against the fully adjudicated human reference set. Confidence intervals are percentile intervals from 10,000 bootstrap resamples clustered by underlying benchmark item. Class labels are abbreviated as timely (tim), superseded (sup), and unclear (unc).}
\label{tab:human-validation}
\end{table*}

\begin{table}[t]
\centering
\small
\begin{tabular}{lrrr}
\toprule
& \multicolumn{3}{c}{Automatic-judge label} \\
\cmidrule(lr){2-4}
Human reference label & Superseded & Unclear & Timely \\
\midrule
Superseded & 67 & 4 & 1 \\
Unclear    & 0  & 23 & 8 \\
Timely & 1  & 1  & 95 \\
\bottomrule
\end{tabular}
\caption{Confusion matrix for all 200 adjudicated responses. Rows are human reference labels and columns are automatic-judge predictions.}
\label{tab:human-validation-confusion}
\end{table}

\section{Per-Model Staleness Results}
\label{app:staleness}

Table~\ref{tab:staleness} reports the within-model recency analysis corresponding to Fig.~2a in the main paper. For each model, we compare accuracy on settled and recent reversals and report the accuracy gap and Fisher's exact-test $p$-value. Statistical significance is assessed after Benjamini--Hochberg correction across the 12 model comparisons; 10 of 12 remain significant.

\begin{table*}[t]
\centering\small
\begin{tabular}{lcccc}
\toprule
Model & Settled & Recent (95\% CI) & Gap & Fisher $p$ \\
\midrule
Qwen2.5-7B & 0.97 & 0.61 [0.50,0.70] & +0.36 & 0.0000*** \\
Llama-3.1-8B & 1.00 & 0.64 [0.54,0.74] & +0.36 & 0.0000*** \\
Qwen2.5-72B & 0.97 & 0.70 [0.60,0.79] & +0.27 & 0.0011** \\
Llama-3.1-70B & 1.00 & 0.74 [0.63,0.82] & +0.26 & 0.0004*** \\
GPT-4o & 1.00 & 0.76 [0.66,0.84] & +0.24 & 0.0008*** \\
MedGemma-27B & 1.00 & 0.76 [0.66,0.84] & +0.24 & 0.0008*** \\
MedGemma-4B & 0.88 & 0.68 [0.57,0.77] & +0.20 & 0.0363* \\
GPT-4.1 & 1.00 & 0.81 [0.71,0.87] & +0.20 & 0.0032** \\
DeepSeek-R1 & 0.94 & 0.76 [0.66,0.84] & +0.18 & 0.0354* \\
Claude-Opus-4.5 & 1.00 & 0.87 [0.79,0.93] & +0.13 & 0.0337* \\
Claude-Sonnet-4.6 & 1.00 & 0.89 [0.80,0.94] & +0.12 & 0.0601 \\
DeepSeek-V3 & 1.00 & 0.90 [0.81,0.94] & +0.10 & 0.0619 \\
\bottomrule
\end{tabular}
\caption{Settled versus recent accuracy per model (recent with Wilson 95\%
CI). Because each model is its own control, the gap supports recency as the
main explanation while residual item-difficulty differences remain possible.}
\label{tab:staleness}
\end{table*}

\section{Cross-Domain Poisoning Results}
\label{app:crossdomain}

Table~\ref{tab:crossdomain-poison} gives the open-model domain grid
corresponding to Fig.~2b in the main paper. All cells use the frozen v2 corpora
(100 law, 60 software, 70 policy items). The rates span $0.17$ to $0.91$.

\begin{table*}[t]
\centering\small
\begin{tabular}{lcccc}
\toprule
Model & Medicine & Law & Software/API & Platform policy \\
\midrule
Qwen2.5-7B  & 0.765 (39/51) & 0.618 (34/55) & 0.354 (17/48) & 0.541 (20/37) \\
Qwen2.5-72B & 0.855 (47/55) & 0.611 (33/54) & 0.250 (12/48) & 0.410 (16/39) \\
Llama-3.1-8B  & 0.661 (37/56) & 0.200 (12/60) & 0.289 (11/38) & 0.417 (20/48) \\
Llama-3.1-70B & 0.906 (58/64) & 0.347 (26/75) & 0.167 (7/42) & 0.412 (21/51) \\
\bottomrule
\end{tabular}
\caption{Poisoning rate by model and domain on the frozen v2 corpora.
Parentheses give flipped over items answered correctly without retrieval, the
denominator used by Eq.~\ref{eq:poisoning}.}
\label{tab:crossdomain-poison}
\end{table*}

\paragraph{Matched up-to-date evidence.}
To distinguish selective-trust failure from a general inability to use
retrieval, we replace the outdated document with its matched up-to-date
counterpart on all 87 medical items. No previously correct answer becomes
incorrect. After retrieval, four models give the timely answer on all 87
items; Qwen-7B and Llama-8B each give it on 86/87. Among items not answered
correctly without retrieval, the up-to-date document recovers 97--100\%
(Table~\ref{tab:current-evidence}).

\begin{table}[t]
\centering\scriptsize
\begin{tabular}{@{}lcc@{}}
\toprule
Model & Timely & Error recovery \\
\midrule
Qwen-7B & 86/87 & 35/36 \\
Qwen-14B & 87/87 & 34/34 \\
Qwen-32B & 87/87 & 28/28 \\
Qwen-72B & 87/87 & 32/32 \\
Llama-8B & 86/87 & 32/33 \\
Llama-70B & 87/87 & 23/23 \\
\bottomrule
\end{tabular}
\caption{Responses after matched up-to-date evidence under the instructed
retrieval prompt. Recovery is computed only over items not answered correctly
without retrieval.}
\label{tab:current-evidence}
\end{table}

\section{Temporal-Applicability Control}
\label{app:temporal-applicability}

The temporal-applicability manifest contains 50 independently verified
transitions: the frozen 21-item set, a 12-item replication, and a 17-item
extension. Every item passed source, validity-boundary, and answer-pair
verification. Automated validation confirms exactly 50 unique items and prompt
invariance: within each valid/stale pair, the evidence, question, options, and
instructions are byte-identical, and only the evaluation date changes.

Each model completes 600 cells: two evaluation dates crossed with date-only,
prose-boundary, and table-boundary formats, with neutral and explicit-follow
instruction variants. The main analysis uses the neutral variants. Answer
options are deterministically randomized, and the constrained two-line output
is parsed without a free-text judge. Across all 2,400 cells, no answer, status,
or label field is empty. Qwen-7B, Qwen-72B, and Llama-70B have no unclear
cells; Llama-8B has 47 malformed cells confined primarily to follow variants,
which remain unclear under the frozen parser.

For each format, the applicability gap is the old-answer rate while the
evidence is valid minus its rate after supersession. Table~\ref{tab:temporal-gap}
reports item-clustered bootstrap 95\% confidence intervals. Qwen-72B makes all
50 required old-to-current transitions under both explicit formats. Llama-70B
makes 47/50 under prose and 50/50 under the table format; the corresponding
date-only counts are 6/50 and 7/50.

\begin{table*}[t]
\centering\small
\begin{tabular}{@{}lccc@{}}
\toprule
Model & Date only & Prose boundary & Table boundary \\
\midrule
Qwen-7B & $.06\ [-.02,.14]$ & $.18\ [.08,.30]$ & $.42\ [.28,.56]$ \\
Llama-8B & $.04\ [.00,.10]$ & $.30\ [.18,.42]$ & $.28\ [.16,.42]$ \\
Qwen-72B & $.12\ [.04,.22]$ & $1.00\ [1.00,1.00]$ & $1.00\ [1.00,1.00]$ \\
Llama-70B & $.14\ [.06,.24]$ & $.94\ [.86,1.00]$ & $1.00\ [1.00,1.00]$ \\
\bottomrule
\end{tabular}
\caption{Applicability gaps on 50 verified transitions under neutral
instructions. Brackets give item-clustered bootstrap 95\% confidence
intervals. Higher values indicate stronger discrimination between dates on
which the same evidence is valid and superseded.}
\label{tab:temporal-gap}
\end{table*}

\section{Evaluation-Date Causal Interventions}
\label{app:temporal-causal}

The causal analysis uses an answer-logit task and retains, for each large
model, up to 20 items that independently reproduce an old-answer preference at
the valid date and a current-answer preference at the stale date. Qwen-72B has
20 items spanning 18 event clusters; Llama-70B has 20 items spanning 20
clusters. These eligibility criteria make the estimates conditional
mechanistic effects rather than all-item behavioral effects.

First, we patch the complete residual state at the evaluation-date span from
the valid prompt into the otherwise identical stale prompt. The transfer is
bidirectional: the reverse intervention moves the valid prompt toward the stale
answer by comparable magnitude. A receiver-to-receiver self-patch is the sham
control. A matched control patches the same local span around the unchanged
historical source date.

At layer 0, the valid-date state transfers $23.08$ logits in Qwen-72B (95\%
cluster-bootstrap CI $[21.06,25.53]$) and $7.82$ in Llama-70B
($[6.92,8.65]$), compared with unpatched date effects of $23.07$ and $7.78$.
The matched source-date shifts are $.04$ and $.03$, and self-patches are zero.

We then trace the transplanted date state into the final prompt-token decision
representation. For Qwen-72B, the effect reaches $3.66$ logits at layer 55
($[3.09,4.40]$), 54\% of the donor effect at layer 59, 80\% at layer 63,
and 90\% at layer 75. For Llama-70B, it reaches $1.21$ at layer 31
($[.99,1.44]$), 75\% at layer 39, and 86\% at layer 71. Matched and sham
controls remain near zero. This supports a propagation-then-integration account
but does not by itself identify a dedicated temporal circuit.

\section{Component, Head, and Specificity Analyses}
\label{app:temporal-components}

\paragraph{Attention and MLP components.}
Within the model-specific integration windows, we separately patch the donor
attention output and MLP output at the final prompt token. The substantive
onset is attention-led in both architectures (Table~\ref{tab:temporal-components}).
Later layers contain alternating attention and MLP contributions, including
negative interventions, so component effects are not additive.

\begin{table*}[t]
\centering\small
\begin{tabular}{@{}lccc@{}}
\toprule
Model/layer & Attention shift [95\% CI] & MLP shift [95\% CI] & Full residual trace \\
\midrule
Llama-70B L31 & $.91\ [.78,1.06]$ & $-.17\ [-.23,-.11]$ & $1.21$ \\
Llama-70B L39 & $2.51\ [2.08,2.97]$ & $.03\ [-.06,.12]$ & $5.86$ \\
Qwen-72B L55 & $2.34\ [2.01,2.74]$ & $-.04\ [-.19,.07]$ & $3.66$ \\
Qwen-72B L59 & $.73\ [.58,.87]$ & $1.66\ [1.28,2.02]$ & $12.45$ \\
\bottomrule
\end{tabular}
\caption{Attention- and MLP-output interventions in the date-to-decision
integration windows. Confidence intervals cluster by the underlying event.}
\label{tab:temporal-components}
\end{table*}

\paragraph{Split-sample head localization.}
All heads are screened on 10 discovery items. The top five positive heads per
layer are frozen and evaluated on 10 untouched confirmation items. Exact
sign-flip tests are Holm-corrected within model. Eight of ten Qwen heads and
seven of fifteen Llama heads remain significant. The frozen Qwen groups shift
the margin by $1.78$ logits at layer 55 (95\% CI $[1.40,2.30]$) and $1.75$
at layer 57 ($[1.26,2.26]$). Llama groups shift it by $1.05$ at layer 31
($[.80,1.31]$), $1.21$ at layer 35 ($[.70,1.81]$), and $3.18$ at layer 39
($[2.31,4.05]$). Matched group controls range from $-.03$ to $.04$ logits;
sham effects are zero.

\paragraph{Frozen-head specificity controls.}
Without reselecting heads or layers, we test the same frozen groups on ordinary
date comparison without retrieved evidence and on a non-temporal numeric
threshold decision. Source-variable donor patches reproduce the full task
effects, while matched-variable controls and self-patches are zero.

\begin{table*}[t]
\centering\small
\begin{tabular}{@{}lccc@{}}
\toprule
Model/layer & Temporal applicability & Date comparison & Numeric threshold \\
\midrule
Qwen-72B L55 & $1.78$ (7.7\%) & $1.12$ (6.2\%) & $.66$ (1.3\%) \\
Qwen-72B L57 & $1.75$ (7.5\%) & $.69$ (3.8\%) & $.27$ (.5\%) \\
Llama-70B L31 & $1.05$ (12.8\%) & $1.39$ (7.6\%) & $1.03$ (4.2\%) \\
Llama-70B L35 & $1.21$ (14.8\%) & $1.44$ (7.8\%) & $1.31$ (5.4\%) \\
Llama-70B L39 & $3.18$ (38.7\%) & $6.69$ (36.4\%) & $8.54$ (34.9\%) \\
\bottomrule
\end{tabular}
\caption{Frozen selected-head-group shifts on 10 held-out items per model.
Parentheses give the group shift as a fraction of the task's mean total
effect. All displayed selected-group bootstrap intervals exclude zero.}
\label{tab:temporal-specificity}
\end{table*}

The specificity results reject a strong dedicated-temporal-gate account.
Llama layer 39 transfers similar fractions of temporal applicability, ordinary
date comparison, and numeric threshold decisions. Earlier Llama groups and the
Qwen groups show greater temporal enrichment but remain active on the controls.
The supported conclusion is a shared comparison-and-decision route with
architecture- and layer-dependent temporal enrichment, not a uniquely temporal
module.

\section{Probe and Logit-Lens Detail}
\label{app:probe}

The probe is a five-fold cross-validated linear classifier on the
decision-token residual (the last prompt token, read before any answer text
exists). Its target is the model's unaided response label, timely or
superseded, not an explicit document-date judgment. Under the matched prompt,
the hidden-state AUROC exceeds a text-only surface baseline for Qwen-7B in
medicine ($0.631$ versus $0.411$, $n=75$), software ($0.714$ versus $0.571$,
$n=17$), and policy ($0.850$ versus $0.400$, $n=12$), but not law ($0.316$
versus $0.461$, $n=36$). Llama-8B provides a medical replication ($0.674$
versus $0.549$, $n=81$). These results support prediction of the unaided
response from pre-generation activations; they do not by themselves establish
explicit recognition of which answer is timely.

The prompt-matched logit lens compares identical continuations with and without
the outdated document. For Qwen-7B, the mean poison-minus-clean answer-margin shift
is $-0.008$ in the early half and $-1.380$ in the late half; for Llama-8B it is
$-0.169$ and $-1.897$. The absolute clean margin favors the timely
continuation through most layers in Qwen but not in Llama. The replicated claim
is therefore the concentration of the matched suppressive effect in later
layers, not an absolute timely-answer preference throughout both networks.

\section{Instruction Gradient and Base-versus-Instruct Detail}
\label{app:gradient}

The gradient uses only items answered correctly without retrieval in every
condition. For Llama-8B, the common set has $53$ items and poisoning rates of
$0.302$, $0.340$, $0.453$, and $0.660$ from organic through instructed
context. For Qwen-7B, the common set has $50$ items and rates of $0.360$,
$0.600$, $0.680$, and $0.760$. Qwen's document-only and instructed endpoints
differ by exact McNemar $p=0.0215$.

Table~\ref{tab:gradient-crossdomain} extends the same design to the frozen
nonmedical corpora. Instructed poisoning exceeds document-only poisoning in
all eight model-domain cells. The intermediate arms are not monotone in every
domain, and only three individual endpoint tests reach $p<0.05$. Pooling the
nonmedical discordant pairs within family gives $28$ increases versus $12$
decreases for Qwen ($p=0.0166$) and $23$ versus $6$ for Llama ($p=0.0023$).

\begin{table*}[t]
\centering\small
\begin{tabular}{llrrrrrc}
\toprule
Model & Domain & $n$ & Organic & Document only & Consider & Instructed & Exact $p$ \\
\midrule
Qwen-7B & Medicine & 50 & .360 & .600 & .680 & .760 & .0215 \\
Qwen-7B & Law & 55 & .436 & .473 & .600 & .618 & .1153 \\
Qwen-7B & Software/API & 48 & .292 & .271 & .250 & .354 & .3438 \\
Qwen-7B & Policy & 37 & .405 & .432 & .595 & .541 & .3438 \\
\addlinespace
Llama-8B & Medicine & 53 & .302 & .340 & .453 & .660 & $1.53{\times}10^{-5}$ \\
Llama-8B & Law & 60 & .167 & .167 & .233 & .200 & .5000 \\
Llama-8B & Software/API & 38 & .158 & .132 & .158 & .290 & .1094 \\
Llama-8B & Policy & 48 & .146 & .229 & .354 & .417 & .0490 \\
\bottomrule
\end{tabular}
\caption{Instruction gradient on items answered correctly without retrieval in
all four conditions. The final column is the exact paired McNemar test for
document-only versus instructed context.}
\label{tab:gradient-crossdomain}
\end{table*}

On those same items, the Llama late-layer shift magnitudes are $1.815$, $1.755$,
$1.783$, and $1.848$; the Qwen values are $1.214$, $1.256$, $1.335$, and
$1.406$. We report the maximum minus minimum across the four conditions as a
fraction of the organic magnitude: $5.2\%$ for Llama and $15.8\%$ for Qwen.
The corresponding behavioral endpoint increases are $119\%$ and $111\%$.
Thus the measured shift is not invariant, but its range is much smaller than
the behavioral change in both families.

To put behavior and the readout on one statistical scale, let $Y_{ic}=1$ when
item $i$ is poisoned under condition $c$, and let $S_{ic}$ be the magnitude of
its matched late-layer shift. We fit separate logistic models for Qwen and
Llama,
\begin{equation}
 \operatorname{logit}\Pr(Y_{ic}=1)=
 \alpha+\beta_c+\gamma(S_{ic}-\bar S_i)+\delta\bar S_i,
 \label{eq:dissociation-model}
\end{equation}
with organic retrieval as the reference condition. Standard errors use a CR1
sandwich covariance clustered by item. Confidence intervals in
Table~\ref{tab:dissociation} come from $2{,}000$ item-level bootstrap
samples.
For Llama, the adjusted instructed-versus-organic odds ratio is $4.63$ (95\%
bootstrap CI $[2.72,9.72]$), while the within-item shift odds ratio per standard
deviation is $0.94$ ($[0.73,1.21]$). The corresponding Qwen estimates are
$4.98$ ($[2.69,12.15]$) and $1.11$ ($[0.85,1.47]$). Joint item-clustered Wald
tests reject no condition effect in Llama ($p=1.6\times10^{-5}$) and Qwen
($p=2.7\times10^{-5}$).

\begin{table}[t]
\centering
\small
\begin{tabular}{@{}lcc@{}}
\toprule
 & Llama-8B & Qwen-7B \\
\midrule
Items/observations & $53/212$ & $50/200$ \\
Strongest/neutral OR & $4.63\ [2.72,9.72]$ & $4.98\ [2.69,12.15]$ \\
Late-layer shift OR & $0.94\ [0.73,1.21]$ & $1.11\ [0.85,1.47]$ \\
Overall instruction $p$ & $1.6\!\times\!10^{-5}$ & $2.7\!\times\!10^{-5}$ \\
\bottomrule
\end{tabular}
\caption{Stronger instructions predict poisoning even after accounting for the measured late-layer shift. Each item is evaluated under four instruction conditions. The instruction odds ratio compares the strongest instruction (``follow the dated documentation'') with neutral retrieval. An OR above 1 indicates greater odds of poisoning. The late-layer OR measures whether changes in the internal readout predict poisoning. Brackets show 95\% bootstrap confidence intervals; $p$ tests the overall effect of instruction condition.}
\label{tab:dissociation}
\end{table}

The primary outcome follows the paper's poisoning definition and counts any
response not judged timely. As a sensitivity analysis, we remove unclear
responses and compare superseded with timely answers. The instructed odds
ratios remain $4.76$ for Llama (95\% bootstrap CI $[2.83,10.60]$) and $4.98$
for Qwen ($[2.72,12.32]$); the within-item shift estimates remain close to one
at $0.96$ ($[0.74,1.26]$) and $1.11$ ($[0.85,1.45]$), respectively.

The older base-versus-instruct $2\times2$ comparison is retained only as an
exploratory analysis. Its instruct-model lens did not use the behavior-matched
chat and answer formatting used in the main comparisons. Behavior is also
item-set sensitive: on the $36$ items answered correctly without retrieval in
all four arms, the Qwen
difference does not survive ($0.56$ versus $0.47$, exact McNemar $p=0.58$),
while Llama reverses direction ($0.61$ versus $0.39$, $p=0.02$). We draw no
main-text conclusion about whether post-training creates the vulnerability.

\section{Factorial Prompt and Status-Judgment Analyses}
\label{app:factorial-recognition}

The prompt factorial crosses two binary features: a neutral dated header versus
the ``most current'' header, and no directive versus the explicit instruction
to follow the document when it conflicts with prior knowledge. Analysis is
restricted to items answered correctly without retrieval in all four cells,
giving 51 Qwen items and 53 Llama items. Table~\ref{tab:factorial-prompt}
reports the paired cell rates.

\begin{table}[t]
\centering\small
\begin{tabular}{lcc}
\toprule
Condition & Qwen-7B & Llama-8B \\
\midrule
Neutral header, no directive & .373 & .302 \\
``Most current,'' no directive & .608 & .340 \\
Neutral header, follow directive & .745 & .660 \\
``Most current,'' follow directive & .765 & .660 \\
\bottomrule
\end{tabular}
\caption{Poisoning rates in the $2\times2$ prompt factorial. Denominators are
$n=51$ for Qwen and $n=53$ for Llama.}
\label{tab:factorial-prompt}
\end{table}

Under the neutral header, the follow directive has odds ratio $4.92$ for Qwen
(clustered 95\% CI $[2.59,9.34]$, $p=1.1\times10^{-6}$) and $4.50$ for Llama
($[2.46,8.23]$, $p=1.1\times10^{-6}$). In both families, 19 discordant items
deteriorate and none improves (exact McNemar $p=3.8\times10^{-6}$). Without
the directive, the ``most current'' header increases Qwen poisoning
(OR $2.61$, $[1.58,4.31]$, $p=1.8\times10^{-4}$) but not Llama poisoning
(OR $1.19$, $[0.84,1.68]$, $p=0.32$). With the directive present, the header
effect is not significant in Qwen (OR $1.11$, $[0.64,1.94]$, $p=0.71$) or
Llama (OR $1.00$, $[0.62,1.61]$, $p=1.00$). Ten thousand item-bootstrap
samples give the same conclusions.

The status-judgment experiment presents the timely and superseded recommendation
from each of the 87 medical items in separate trials. The prompt supplies the
source and evaluation dates and requests one of three labels: current,
superseded, or uncertain. Generation is greedy; all 348 responses satisfy the
label parser. Table~\ref{tab:recognition-action} reports the full label counts
and the answering outcome on the factorial subset answered correctly without
retrieval.

\begin{table}[t]
\centering\small
\begin{tabular}{@{}llrrr@{}}
\toprule
Model & Document & C & S & U \\
\midrule
Qwen-7B & Up-to-date & 86 & 0 & 1 \\
 & Outdated & 10 & 1 & 76 \\
Llama-8B & Up-to-date & 48 & 39 & 0 \\
 & Outdated & 0 & 87 & 0 \\
\bottomrule
\end{tabular}
\medskip

\begin{tabular}{@{}llrrr@{}}
\toprule
Model & Warning & $n$ & Failure & Superseded \\
\midrule
Qwen-7B & Uncertain & 49 & 36 & 33 \\
Llama-8B & Superseded & 53 & 35 & 32 \\
\bottomrule
\end{tabular}
\caption{Status judgments and subsequent behavior under the neutral header
plus follow directive. C/S/U denotes the prompt labels ``current''
(up-to-date), superseded, and uncertain. ``Failure'' includes superseded and
unclear answers; ``exact superseded'' includes
only answers judged to express the superseded recommendation. The judgments and
answers come from separate prompts using the same dated claim.}
\label{tab:recognition-action}
\end{table}

The confusion counts rule out a calibrated-recognition interpretation. Qwen
usually treats outdated evidence as uncertain rather than identifying the actual
supersession. Llama labels every outdated-document arm superseded but also overcalls
supersession for 39 up-to-date documents. The paired result therefore concerns whether an explicit
status warning governs later answering behavior; it does not establish that
the model encodes the specific historical replacement.

\section{Steering and Patching Detail}
\label{app:steering}

The matched Qwen experiment separates direction learning from evaluation. It
uses 20 training and 14 held-out medical flips, and 17 training and 14 held-out
law flips. At each $\alpha\in\{0.25,0.5,1,2\}$, the learned direction is
compared with 20 equal-norm random directions on the same held-out items. The
best medical rescue is $2/14$ at $\alpha=0.25$ and $0.5$; the best law rescue
is $1/14$ at $\alpha=0.25$. Neither exceeds the random-direction 95th
percentile at any magnitude. The positive control changes $11/14$ medical and
$12/14$ law outputs, so the intervention setup can alter generation.

The older Llama medical run reports a peak $27\%$ rescue but used one random
direction only at $\alpha=2.0$. It remains exploratory and is not used for a
directional-versus-random claim. The prompt-matched patching run evaluates 39
Qwen and 37 Llama flips. All-band clean-state patching makes the timely
continuation preferred in $1/39$ Qwen and $2/37$ Llama cases; random-item late
patching does so in $0/39$ and $1/37$. The mean teacher-forced margin
restoration is $0.307$ for Qwen and $0.168$ for Llama, compared with $0.260$
and $0.155$ for the random-item late control.

For steering, the intervention site is $\lfloor0.6L\rfloor$. The
timely-answer direction is the normalized difference between mean
timely- and superseded-statement residuals, and the added vector is scaled by the mean
residual norm. We evaluate $\alpha\in\{0,0.25,0.5,1,2\}$; $\alpha=0$ checks
that the original flip is reproduced, and equal-norm random unit vectors are
the negative control. For patching, early and late denote the first and second
halves of the decoder stack. We replace only the last-prompt-token residual
and score the teacher-forced mean log-probability margin between the timely
and superseded continuations. The control patches the late band from a random
other item's clean run.

\section{Geometry Decomposition}
\label{app:geometry}

The poison-induced decision-token shift is decomposed into a radial component
along an unaided-correctness direction and an orthogonal remainder. At the late layer
the instructed orthogonal-to-radial ratio is $6.63$ ($8.50$ at the mid layer).
This ratio alone is not evidence of a special rotation in a high-dimensional
space. Against $1{,}000$ random axes, the observed mean absolute projection
onto the unaided-correctness axis is larger, not smaller: $21.231$ versus $1.719$ late and
$1.349$ versus $0.152$ mid. A benign up-to-date document also produces similar
ratios ($7.14$ late and $7.32$ mid). The controlled analysis therefore does not
support a simple rotation account, and we do not use it to explain the steering
or patching results.

Concretely, let $\hat{u}$ be the normalized difference between the mean clean
residuals for items answered correctly and incorrectly without retrieval, and let
$\Delta=h_{\mathrm{poison}}-h_{\mathrm{clean}}$. We report the signed axial
shift $\Delta_{\parallel}=\Delta^{\top}\hat{u}$ and orthogonal magnitude
$\Delta_{\perp}=\|\Delta-\Delta_{\parallel}\hat{u}\|_2$ at
$\lfloor0.6L\rfloor$ and $\lfloor0.9L\rfloor$. These labels describe a
decomposition relative to the measured unaided-correctness axis; they do not identify
independent causal circuits.

\section{Retrieval Gate and Stress Test}
\label{app:gate}

A first, idealized experiment establishes the proof of principle. A recency
rule drives poisoning from $16/66$ to $0/66$ for GPT-4o and from $27/77$ to
$0/77$ for Claude (exact McNemar $p\le6\times10^{-5}$) when the up-to-date
document carries correct date metadata. Those zero-poisoning results are not
the realistic defense estimate.

The retrieval-defense evaluation uses the frozen 317-item benchmark and an
index with one up-to-date and two outdated documents per item. It evaluates
correct, absent, noisy, and
incorrect date metadata. Both dense systems use
\texttt{sentence-transformers/all-MiniLM-L6-v2}. Plain dense retrieval returns
the top semantic match. The re-ranker takes the top four and scores document
$d$ as
\begin{equation}
s(d,q)=0.5\,\cos(d,q)+0.5\,r(d)+0.1\,I_{\mathrm{sup}}(d),
\end{equation}
where $r(d)$ is the document year min--max normalized within the index and
$I_{\mathrm{sup}}$ marks explicit supersession language. It also
flags near-tied semantic matches with conflicting dates. This fixed rule is
used throughout this evaluation.

The re-ranker selects the up-to-date document more often than dense retrieval in
8 of 12 non-clean cells, but it is not uniformly safe. Under incorrect dates,
up-to-date-document selection falls to zero in every domain, below the dense
baseline. Downstream generation was evaluated in law and medicine with Qwen-7B
and GPT-4o. With correct dates, re-ranking lowers poisoning in all four cells
by $4.6$ to $10.0$ percentage points; three exact paired tests reach $p<0.05$.
With dates absent, three cells improve by $1.0$ to $4.0$ points, none
significantly, and one worsens by $1.1$ points
(Table~\ref{tab:stage8-downstream}).

\begin{table}[t]
\centering\small
\begin{tabular}{@{}llcc@{}}
\toprule
Domain & Model & Correct dates & Dates absent \\
\midrule
Law & Qwen-7B & \shortstack{$.640\to.540$\\$p=.0129$} &
  \shortstack{$.670\to.660$\\$p=1.000$} \\
Law & GPT-4o & \shortstack{$.700\to.610$\\$p=.0225$} &
  \shortstack{$.750\to.710$\\$p=.125$} \\
Medicine & Qwen-7B & \shortstack{$.609\to.563$\\$p=.125$} &
  \shortstack{$.598\to.575$\\$p=.500$} \\
Medicine & GPT-4o & \shortstack{$.621\to.529$\\$p=.0215$} &
  \shortstack{$.494\to.506$\\$p=1.000$} \\
\bottomrule
\end{tabular}
\caption{Downstream poisoning in the retrieval-defense evaluation under dense
retrieval and recency re-ranking (dense $\to$ re-ranked). Exact paired McNemar
tests use the retained item-level verdicts.}
\label{tab:stage8-downstream}
\end{table}

\section{Frontier-Model Cells}
\label{app:frontier}

GPT-5.5 unaided accuracy is $0.92$ to $1.00$ across domains. Poisoning: medicine
$0.89$ ($74/83$, Wilson $[0.81, 0.94]$); code $0.11$ ($2/18$); policy $0.13$
($2/15$); law $0.26$ ($12/46$), the non-medicine cells small and
statistically overlapping (Wilson intervals spanning roughly $0.03$ to
$0.40$; code and policy ran on the smaller pre-expansion item set).

\section{Reproducibility}
\label{app:repro}

The release contains the benchmark items with official-source provenance and
archived snapshots, the poisoning corpora, the audit-poison-probe-defense
code, per-run outputs (\texttt{results/*.json}, from which every number in
this paper is generated), figure scripts, and the pinned model revisions of
Appendix~\ref{app:models}.
The behavioral panel is produced by \texttt{poison\_forced.py}; probe and
logit-lens outputs by \texttt{dtoken\_probe.py},
\texttt{dtoken\_layer\_sweep.py}, and \texttt{logit\_lens\_poison.py}; causal
interventions by \texttt{steer\_poison.py} and
\texttt{causal\_patch\_poison.py}; the item-clustered dissociation test by
\texttt{analyze\_instruction\_dissociation.py}; geometry by
\texttt{geom\_decompose.py}; the prompt factorial and status-judgment analysis
by \texttt{analyze\_elite\_paper.py}; and the retrieval experiments by
\texttt{eval\_guard.py}, \texttt{eval\_guard\_stress.py}, and
\texttt{stage8\_retrieval\_defense.py}. The temporal-applicability panel is
produced by \texttt{temporal\_applicability\_choice.py} and
\texttt{analyze\_temporal\_applicability\_choice.py}; its causal analyses by
\texttt{temporal\_date\_activation\_patch.py},
\texttt{temporal\_date\_causal\_trace.py},
\texttt{temporal\_date\_component\_trace.py},
\texttt{temporal\_date\_head\_trace.py}, and
\texttt{temporal\_head\_specificity\_control.py}, with the correspondingly
named \texttt{analyze\_*} scripts. The intervention
and retrieval scripts fix their sampling seed to zero. Environment manifests
are provided in \texttt{requirements.txt} and \texttt{pyproject.toml}. Existing
behavioral and intervention aggregates retain item-level labels and generated
answers. The retrieval-defense evaluation retains $1{,}496$ item-level
retrieval decisions, answers, and judge verdicts in
\texttt{results/stage8\_downstream\_rows.jsonl}; the paired
statistics are generated by \texttt{analyze\_stage8\_downstream.py}.

Table~\ref{tab:repro-settings} records the final experimental settings and the
only development sweeps. The steering coefficients were evaluated as a fixed
grid and are all reported; no coefficient was selected after observing the
outcome. The decision-token layer sweep evaluated every decoder layer and used
cross-validated AUROC only to identify the best readout layer. Retrieval weights
and all remaining settings were specified before the corresponding evaluation
and were not tuned on test outcomes.
Unless otherwise stated, each model--item--condition combination was evaluated
once using deterministic or temperature-zero decoding; uncertainty estimates
resample items rather than repeated generations.

\begin{table*}[t]
\centering
\small
\begin{tabular}{@{}p{0.19\textwidth}p{0.34\textwidth}p{0.39\textwidth}@{}}
\toprule
Component & Final setting & Values tried or selection rule \\
\midrule
Behavioral generation & Temperature 0 for hosted models; greedy decoding for open-weight models; output limits 220 and 200 tokens, respectively & Fixed settings; no decoding sweep \\
Automatic judge & GPT-4o; three labels; option order set by stable item--condition hash; four-token output limit & Fixed protocol; no judge prompt selected on the evaluation outcomes \\
Decision-token probe & Last prompt token; standardized logistic regression, $C=0.5$; five-fold cross-validation; layer $\lfloor0.6L\rfloor$ & Layer-sweep analysis evaluates all $L$ decoder layers; maximum cross-validated AUROC identifies the descriptive best layer \\
Surface baseline & TF--IDF, 1--2 grams, at most 4,000 features; logistic regression, $C=1$ & Fixed matched baseline \\
Steering intervention & Layer $\lfloor0.6L\rfloor$; 50/50 train--held-out split; held-out cap 14; 20 random directions & $\alpha\in\{0,0.25,0.5,1,2\}$, with $\alpha=4$ as a positive control; all values reported \\
Causal patching & Decision-token residual; early, late, all-layer, and random late-layer controls & Early/late split fixed at $L/2$; every planned band reported \\
Temporal applicability & 50 independently verified items; date-only, prose-boundary, and table-boundary formats; neutral and follow-up prompts & Fixed answer-option order and parser; prompt-invariance requirement; item-cluster bootstrap intervals \\
Date-state causal trace & Up to 20 behaviorally eligible items per model; bidirectional evaluation-date-span patching; source-date and self-patch controls & Decision layers selected from the complete layer trace; all planned controls reported \\
Temporal head localization & 10 discovery and 10 confirmation items; five strongest positive heads per layer frozen after discovery & Exact sign-flip tests with Holm correction; all-head patching retained as a positive control \\
Temporal specificity controls & Frozen layers and head groups; date-comparison and matched numeric-threshold tasks & No layer, head, or control-task reselection after viewing control outcomes \\
Geometry & Layers $\lfloor0.6L\rfloor$ and $\lfloor0.9L\rfloor$; 1,000 random axes; seed 0 & Fixed before evaluation \\
Retrieval defense & Top four candidates; score $0.5\cos(d,q)+0.5r(d)+0.1I_{\mathrm{sup}}(d)$; corpus seed 7 & Fixed rule; no coefficient search \\
Uncertainty and tests & 10,000 item-cluster bootstrap samples for human validation; Wilson intervals; exact Fisher, McNemar, and Cochran--Mantel--Haenszel tests; Benjamini--Hochberg correction & Prespecified analysis; human-validation bootstrap seed 20260808 \\
\bottomrule
\end{tabular}
\caption{Final settings and development ranges for the reported experiments. $L$ denotes the number of decoder layers.}
\label{tab:repro-settings}
\end{table*}

The principal open-weight batches ran on two NVIDIA B200 GPUs with 180~GB of
memory each; some development runs used Apple Silicon MPS or CPU. Exact CPU
models, operating-system images, wall-clock totals, and historical package
states were not retained for every run. We record this limitation rather than
reconstructing missing metadata.

\end{document}